\documentclass[10pt,journal,compsoc]{IEEEtran}
\usepackage[utf8x]{inputenc}
\pdfoutput=1

\usepackage{graphicx}
\usepackage[cmex10]{amsmath}
\usepackage{amsfonts}
\usepackage{array}
\usepackage{algorithm}
\usepackage{algpseudocode}
\usepackage{multirow}

\ifCLASSOPTIONcompsoc
\else
\fi

\ifCLASSOPTIONcompsoc
  \usepackage{caption}
  \usepackage[font=normalsize,labelfont=sf,textfont=sf]{subfig}
\else
  \usepackage[caption=false]{caption}
  \usepackage[font=footnotesize]{subfig}
\fi
\usepackage[pagebackref=true,breaklinks=true,letterpaper=true,bookmarks=false,pdfborder={0 0 0}]{hyperref}
\hypersetup{
 pdfauthor={Philip G. Lee and Ying Wu},
 pdftitle={Scalable Matting: A Sub-linear Approach}
}
\DeclareMathOperator{\spanop}{span}
\DeclareMathOperator{\nnz}{nnz}
\providecommand{\abs}[1]{\left\lvert #1 \right\rvert}
\providecommand{\norm}[1]{\left\lVert #1 \right\rVert}
\providecommand{\inner}[2]{\langle #1, #2 \rangle}
\providecommand{\vect}[1]{\boldsymbol{#1}}
\providecommand{\defeq}{\stackrel{\Delta}{=}}
\providecommand{\order}[1]{\mathcal{O}\left( #1 \right)}

\title{Scalable Matting: A Sub-linear Approach}
\author{
  Philip~G.~Lee,
  Ying~Wu,~\IEEEmembership{Senior Member,~IEEE}
  \IEEEcompsocitemizethanks{
   \IEEEcompsocthanksitem The authors are with the Department of Electrical
    Engineering and Computer Science, Northwestern University,
    2145 Sheridan Road, Evanston IL 60208.\protect\\
    E-mail: {\tt\small PhilipLee2012@u.northwestern.edu \; yingwu@eecs.northwestern.edu}
  }
  \thanks{}
}

\begin{document}

\IEEEcompsoctitleabstractindextext{
  \begin{abstract}
   Natural image matting, which separates foreground from background, is a very
   important intermediate step in recent computer vision algorithms. However,
   it is severely underconstrained and difficult to solve.
   State-of-the-art approaches include matting by graph Laplacian, which
   significantly improves the underconstrained nature by reducing the solution
   space. However, matting by graph Laplacian is \textit{still} very difficult
   to solve and gets much harder as the image size grows: current iterative
   methods slow down as $\order{n^2}$ in the resolution $n$. This creates
   uncomfortable practical limits on the resolution of images that
   we can matte. Current literature mitigates the problem, but they all remain
   super-linear in complexity. We expose properties of the problem that remain
   heretofore unexploited, demonstrating that an optimization
   technique originally intended to solve PDEs can be adapted to take advantage
   of this knowledge to solve the matting problem, not heuristically, but
   \textit{exactly} and with sub-linear complexity. This makes ours the
   most efficient matting solver currently known by a very wide margin and
   allows matting finally to be practical and scalable in the future as consumer
   photos exceed many dozens of megapixels, and also relieves matting from being
   a bottleneck for vision algorithms that depend on it.
  \end{abstract}
  % Note that keywords are not normally used for peer review papers.
  %\begin{keywords}
  % Computer Society, IEEEtran, journal, \LaTeX, paper, template
  %\end{keywords}
}

\maketitle

% is not selected <OR> if conference mode is selected - because compsoc
% conference papers position the abstract like regular (non-compsoc)
%\IEEEdisplaynotcompsoctitleabstractindextext

% \fi
%\IEEEpeerreviewmaketitle

\section{Introduction}%========================================================

\IEEEPARstart{T}{he} problem of extracting an object from a natural scene is
referred to as alpha matting. Each pixel $i$ is assumed to be a convex combination
of foreground and background colors $F_i$ and $B_i$ with $\alpha_i \in [0,1]$
being the mixing coefficient:
\begin{align}
   I_i = \alpha_i F_i + (1-\alpha_i)B_i.
\end{align}

Besides the motivation to solve the matting problem for graphics purposes like
background replacement, there are many computer vision problems that use
matting as an intermediate step like dehazing \cite{he2009single},
deblurring \cite{dai2008motion}, and even tracking \cite{fan2010closed} to
name a few.

Since there are 3 unknowns to estimate at each pixel, the problem is severely
underconstrained. Most modern algorithms build complex local color or feature
models in order to estimate $\alpha$.
One of the most popular and influential works is \cite{levin2008closed}, which
explored graph Laplacians based on color similarity as an approach to solve
the matting problem.

\begin{figure}
 \centering
 \begin{tabular}{cc}
  \includegraphics[width=3.67cm]{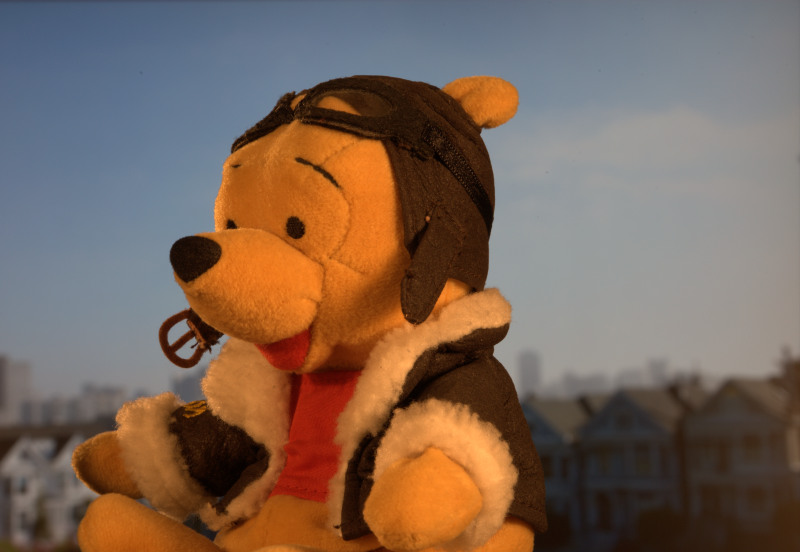} & \includegraphics[width=3.67cm]{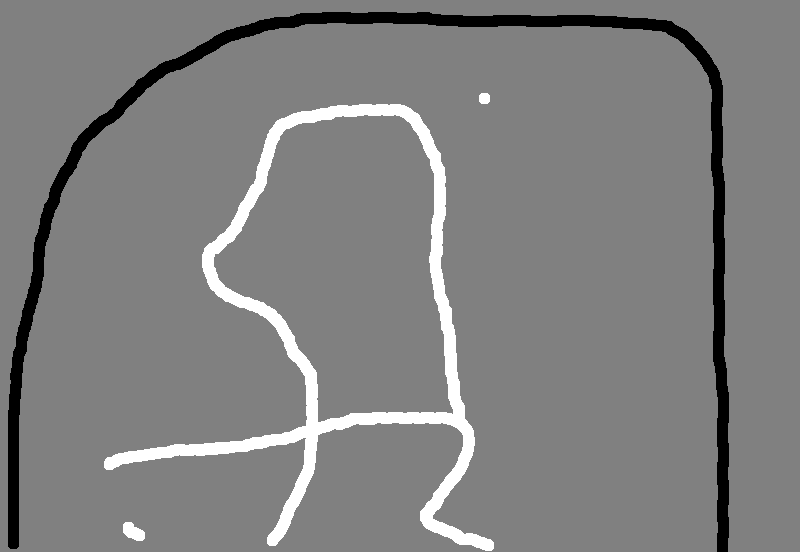} \\
  (a) & (b) \\
  \multicolumn{2}{c}{\includegraphics[width=3.67cm]{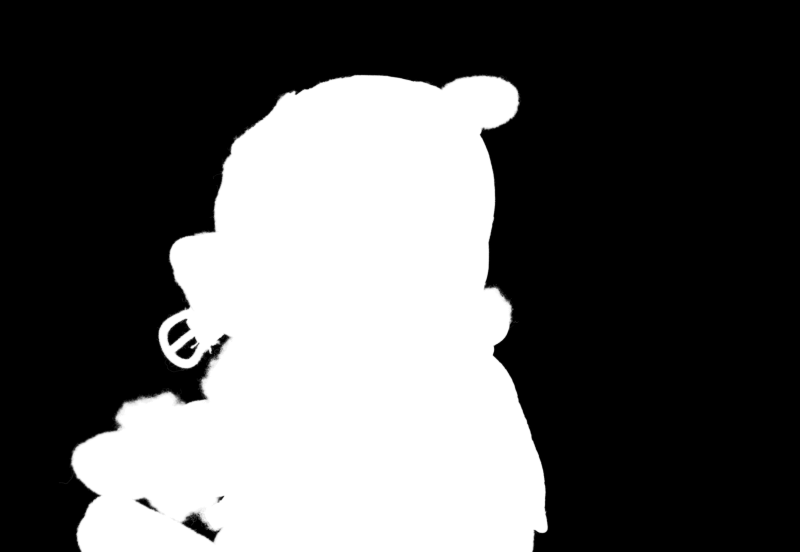}} \\
  \multicolumn{2}{c}{(c)}
 \end{tabular}
 \caption{
  Input image (a). User constraints or ``scribbles'' (b). Desired output matte
  (c).
 }
 \label{fig:theProblem}
\end{figure}

The model that most Laplacian-based matting procedures use is
\begin{align}
 p\left( I \mid \alpha \right) & \propto \exp\left( -\frac{1}{2} \norm{\alpha}^2_L \right) \; \text{and}\\
 \alpha_\text{MLE} &= \arg \max_\alpha p( I \mid \alpha ), \label{eq:mle}
\end{align}
where $L$ is a per-pixel positive semi-definite matrix that represents a graph
Laplacian over the pixels, and $\norm{ \alpha }^2_L = \inner{\alpha}{L\alpha}$ is the induced
norm. Typically $L$ is also very sparse as the underlying graph representing the
pixels is connected only locally to its immediate spatial neighbors, and $L$ is
directly related to the adjacency matrix (see \S\ref{subsec:laplacian}). However, $L$ always
has positive nullity (several vanishing eigenvalues), meaning that the estimate
\eqref{eq:mle} is not unique. So, user input is required to provide a prior to resolve the
ambiguity. A crude prior that is frequently used is:
\begin{align}
 p(\alpha) \propto
   \begin{cases}
    1 & \text{if} \; \alpha \; \text{is consistent with user input} \\
    0 & \text{o.w.}
   \end{cases}.
\end{align}
This is often implemented by asking the user to provide ``scribbles''
to constrain some pixels of the matte to 1 (foreground) or 0 (background) by
using foreground and background brushes to manually paint over parts of the
image. We give an example in Fig. \ref{fig:theProblem}.

To get the constrained matting Laplacian problem into the linear form
$A \alpha = f$, one can choose
\begin{align}
 A &\defeq (L + \gamma C), \; \text{and} \nonumber \\
 f &\defeq \gamma g, \label{eq:linearSystem}
\end{align}
where $C$ is diagonal, $C_{ii}=1$ if pixel $i$ is user-constrained to value
$g_i$ and 0 o.w., and $\gamma$ is the Lagrange multiplier for those constraints.
If there are $n$ pixels in the image $I$, then $A$ is $n \times n$ with
$\nnz(A) \in \order{n}$.

Since $A$ is symmetric and positive definite (with enough
constraints), it is tempting to solve the system by a Cholesky decomposition.
However, this is simply impossible on a modest-size image on modern consumer
hardware, since such decompositions are dense and rapidly exhaust gigabytes of
memory when the resolution is larger than around 10,000 pixels ($316 \times 316$).
The reason is that these decompositions are dense, though the input matrix is
sparse. So, if each entry is 8 B, then the number of GB required to store the
decomposition of a sparse $10,000 \times 10,000$ system is
$8(10,000^2)/(2(10^9)) = 40$ GB. Further, although they are fast for small systems,
the time complexity is $\order{n^3}$, which will eventually dominate, even if we
have unlimited memory.

It is also tempting to incorporate the user-provided equality constraints directly, which reduces
the system size by the number of constrained pixels. In order to solve this
reduced system in memory would still require the number of unconstrained pixels
to be less than about 40k on a system with 8 GB memory. This is also impractical,
because it places the burden on the user or some other automated algorithm to
constrain at least 95\% of a megapixel image, and much more at higher resolutions.

There are a number of methods to \textit{approximately} solve the problem by
relying on good heuristics about natural images and the extracted alpha mattes
themselves. For example, \cite{he2010fast} split the problem into multiple
independent pieces, and \cite{xiao2014fast} perform segmentation to reduce
the problem size.

If we desire to avoid heuristics and obtain an exact solution, then it is
necessary to resort to more memory-efficient iterative algorithms, which
can take advantage of the sparsity of $A$. There are many traditional
iterative methods for solving the full-scale matting problem such as
the Jacobi method, succesive over-relaxation, gradient descent, and conjugate
gradient descent. Although memory-efficient, they are all quite slow,
especially as the resolution increases. There is a fundamental reason for this:
the condition number of the system grows as $\order{n^2}$ in the number of pixels,
making these linear solvers also slow down at something like $\order{n^2}$.

Even the ``fast'' matting methods only improve the complexity by a
constant factor. For example, by splitting the problem into $k$
independently-solved pieces, \cite{he2010fast} changes the complexity to
$\order{ k(n/k)^2 } = \order{ n^2/k } = \order{n^2}$. \cite{xiao2014fast} at first seems
to solve the problem efficiently by segmenting the problem into $m \ll n$
segments where $m$ is constant w.r.t. $n$. However, the complexity is hidden
in performing the segmentation, which is still at least $\order{n^2}$. No matter
what engineering we do to reduce these constant factors, it is the underlying
solvers that prevent us from improving the overall complexity to get an
algorithm that is truly scalable.

Iterative methods can be characterized by their per-iteration error
reduction ratio, which is called the convergence rate. The closer
to 0, the better, and the closer to 1, the worse the method is.
Currently-used iterative methods are bound by the condition number of the
system, and therefore have convergence rates like $1-\order{n^{-2}}$, slowing down dramatically
as the problem gets large (\S \ref{subsec:Relaxation}).

There is a property of mattes of natural images that has not been fully
exploited: they tend to be constant almost everywhere (Fig. \ref{fig:theProblem}.c).
In fact, for opaque foreground objects, the only regions that contribute to the
objective function are the pixels on the edge of the object. Computation on
non-boundary pixels is wasteful. Further, as the
resolution grows, the ratio of the number of edge pixels to the rest of
the pixels goes to zero. This means if we can find a solver that
exploits these properties to focus on the boundary, it should be
\textit{more} efficient as $n$ grows. Since these properties hold independent
of the choice of the Laplacian, such a solver would be general and much more
natural than the heuristics currently applied to the Laplacian itself.

In this paper, we propose the use of multigrid algorithms to efficiently
solve the matting problem by fully exploiting these properties. Multigrid methods are briefly mentioned by
\cite{szeliski2006locally} and \cite{he2010fast} for solving the matting
problem, concluding that the irregularity of the Laplacian is too much for
multigrid methods to overcome. We thoroughly analyze those claims, and find
that conclusion to be premature. This paper shows for the first time a solver
of \textit{sub}-linear time complexity, ($\order{ n^{0.752} }$ average case)
that solves the full-scale problem without heuristics, works on any choice of matting
Laplacian, sacrifices no quality, and finally provides a scalable solution for
natural image matting on ever larger images.

\section{Background}%==========================================================

\subsection{The Matting Laplacian}%--------------------------------------------
\label{subsec:laplacian}

Matting by graph Laplacian is a common technique in matting literature
\cite{levin2008closed,zheng-learning,he2010fast,lee2011nonlocal}. Laplacians
play a significant role in graph theory and have wide applications, so we will take
a moment to describe them here. Given an undirected graph $G = (V,E)$, the adjacency matrix
of $G$ is a matrix $A^G_{ij}$ of size $\lvert V \rvert \times \lvert V \rvert$
that is positive $\forall v_i v_j \in E$. $A^G_{ij}$ is the
``weight'' on the graph between $v_i$ and $v_j$. Further define
$D^G_{ii} = d(v_i) = \sum_{j \neq i} A^G_{ij}$ to be a diagonal matrix containing
the degree of $v_i$, and let $x$ be a real-valued function
$V \rightarrow \mathbb{R}$. The Laplacian $L^G=D^G-A^G$ defines the
quadratic form on the graph
\begin{align}
  q(x) \defeq x^T L^G x = \sum_{v_iv_j \in E} A^G_{ij}(x_i - x_j)^2.
\end{align}
By construction, it is easy to see that $L^G$ is real, symmetric, and
positive semi-definite, and has $\vect{1}$ (the constant vector) as its first
eigenvector with eigenvalue $\lambda_1 = 0$.

This quadratic form plays a crucial role in matting literature, where
the graph structure is the grid of pixels, and the function over the vertices
is the alpha matte $\alpha \in [0,1]^{\lvert V \rvert}$. By designing
the matrix $A^G_{ij}$ such that it is large when we expect $\alpha_i$ and
$\alpha_j$ are the same and small when we expect they are unrelated, then
minimizing $q(\alpha)$ over the alpha matte is finding a matte that
best matches our expectations. The problem is that, as we have shown, one
minimizer is always the constant matte $\vect{1}$, so extra constraints are
always required on $\alpha$.

Many ways have been proposed to generate $A^G$ so that a good matte can be
extracted. In \cite{sun2004poisson}, $A^G$ is constructed to represent simple
4-neighbor adjacency. In \cite{levin2008closed}, $A^G_{ij}$ is given as
{
\small
\begin{align}
  \sum_{k \mid (i,j) \in w_k}
                \frac{1}{\lvert w_k \rvert}
                \left(
                  1 + \inner{I_i - \mu_k}{I_j- \mu_k}_{(\Sigma_k + \epsilon I)^{-1}}
                \right),
                \label{eq:levinAffinity}
\end{align}
}
where $w_k$ is a spatial window of pixel $k$, $\mu_k$ is the mean color in
$w_k$, and $\Sigma_k$ is the sample color channel covariance matrix in $w_k$.

In \cite{zheng-learning}, the color-based affinity in eq. \eqref{eq:levinAffinity}
is extended to an infinite-dimensional feature space by using the kernel trick
with a gaussian kernel to replace the RGB Mahalanobis inner product, and
\cite{lee2011nonlocal} does something similar on patches of colors instead of
single colors, with more flexibility in the model and assumptions.
\cite{duchenne2008segmentation} provides many other uses of
graph Laplacians, not only in matting or segmentation, but also in other
instances of transductive learning.
\cite{singaraju2009new} demonstrated that
\cite{levin2008closed}'s Laplacian intrinsically has a nullity of 4, meaning
that whatever prior is used, it must provide enough information for each
of the 4 subgraphs and their relationships to each other in order to find a
non-trivial solution.

No matter how we choose $A^G$, the objective is to minimize $q(x)$ subject to
some equality constraints (e.g. the user scribbles mentioned above). Although
it is simply a constrained quadratic program, it becomes very difficult to
solve as the system size grows past millions of pixels, and necessitates
investigation to be practical.

\subsection{Relaxation}%-------------------------------------------------------
\label{subsec:Relaxation}

No matter what method we choose to construct the matting Laplacian, we end up
with a large, sparse,
ill-conditioned problem. There are a number of methods to solve large linear systems
$Au=f$ for $u$. Traditional methods include Gauss-Seidel and successive
over-relaxation (SOR). These are \textit{relaxation} methods, where the
iteration is of the affine fixed-point variety:
\begin{align}
 u \leftarrow G(u) \defeq Ru+g.
\end{align}
If $\hat{u}$ is the solution, it must be the case that it is a fixed
point of the iteration: $\hat{u} = R\hat{u}+g$. For analysis, we can define
the error and residual vectors
\begin{align}
 e &\defeq \hat{u}-u \; \text{and}\nonumber \\
 r &\defeq f-Au \nonumber
\end{align}
so that original problem is transformed to $Ae=r$. We can then see that
\begin{align}
 e &\leftarrow Re \; \text{i.e.} \nonumber \\
 e^{\text{new}} &= Re \label{eq:errorIteration}
\end{align}
is the iteration with respect to the error. This tells us that the convergence
rate $\norm{e^\text{new}}/\norm{e}$ is bounded from above by the spectral radius
$\rho(R)$. So, a simple upper bound for the error reduction at iteration $j$ is
$\norm{e^j}/\norm{e^1} \leq \rho(R)^j$. So, for quickest convergence and
smallest error, we want $\rho(R)$ to be much less than 1.

For our kind of problems (discrete Laplace operators), the Jacobi method's
iteration matrix $R_J$ gives a convergence rate of $\rho(R_J) \leq 1-a/n^2$,
where $n$ is the number of variables and $a$ is a constant. The Gauss-Seidel
iteration matrix gives $\rho(R_G) \leq \rho(R_J)^2$, meaning it converges
twice as fast as the Jacobi method. There are similar results for SOR,
gradient descent, and related iterative methods.

Recall $\rho(R) = \max_i \abs{\mu_i}$, where $\mu_i$ are the eigenvalues
of $R$. So, to analyze the convergence, we need to
understand the eigenvalues. We can decompose the error $e$ in the basis of
eigenvectors $v_i$ of $R$:
\begin{align}
 e^1 = \sum_{i=1}^n a_i v_i.
\end{align}
After $j$ iterations, we have
\begin{align}
 e^j = \sum_{i=1}^n \mu_i^j a_i v_i.
\end{align}
So, the closer $\mu_i$ is to zero, the faster the $v_i$ component of $e$
converges to 0. For Laplace operators, we explain
in the example below that $i$ corresponds to a generalization of 
frequency, and $\rho(R) = \mu_1$, the lowest ``frequency''
eigenvalue. 

Since $\mu_1$ is closest to 1, the $v_1$ (low-frequency) component dominates the
convergence rate, while the higher-frequency components are more easily
damped out (for this reason, relaxation is often called
\textit{smoothing}). Further, we explain that the convergence rate rapidly approaches 1 as
$n$ grows. To break this barrier, we need a technique that can handle the
low-frequencies just as well as the high frequencies.

\subsubsection{Motivating Example}%--------------------------------------------

To understand why, let us take the 1D Laplace equation with Dirichlet boundary
conditions as an example from \cite{mccormick1987multigrid}:
\begin{align}
 Au &= f, \; \text{where} \label{eq:1dLaplace} \\
 A &= -\nabla^2, \; \text{and} \nonumber \\
 f &=0 \nonumber \\
 \text{s.t.} \; u_0 &= u_{n+1} = 0. \nonumber
\end{align}
This problem has two fixed variables $u_0$ and $u_{n+1}$, and $n$ free
variables $u_1,\dotsc,u_n$. $A$ is $n \times n$ and is the operator
$-\nabla^2$ corresponding to convolution with the discrete Laplace kernel
$[-1,2,-1]$. The underlying graph in this case has node $i$ adjacent to nodes
$i-1$ and $i+1$. We use this example because it is directly analagous to
matting, but easier to analyze. Notice that the problem has a unique solution
$u=0$.

Without loss of generality, we assume that the $u_i$ are spaced $h=1/(n+1)$
units apart, corresponding to a continuous function $u$ whose domain is the
unit interval $[0,1]$. We write the problem in Eq. \eqref{eq:1dLaplace}
as
\begin{align}
 -u_{i-1}+2u_i-u_{i+1} = 0, \; \forall 1 \leq i \leq n,
\end{align}
which naturally incorporates the constraints, and gives us a symmetric, banded,
positive-definite matrix,
\begin{align}
 A = \left[
 \begin{array}{rrrrr}
   2 & -1 &  0 &  0 & \\
  -1 &  2 & -1 &  0 & \\
   0 & -1 &  2 & -1 & \\
     &    &  \ddots & \ddots & \ddots
 \end{array}
 \right].
\end{align}

\begin{figure}
 \centering
 \begin{tabular}{c}
  \includegraphics[width=8cm]{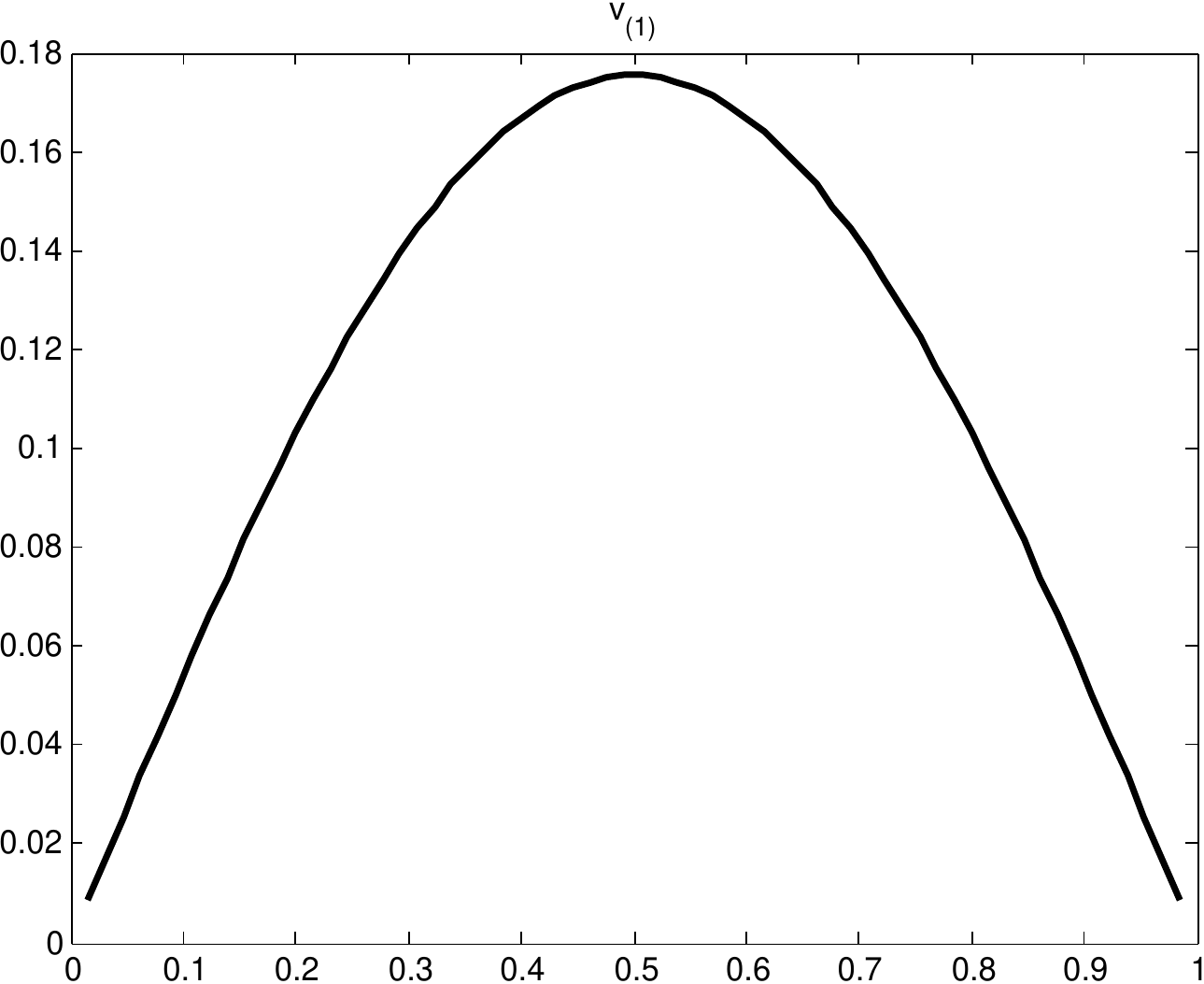}  \\ %& \includegraphics[width=4cm]{images/v64_2.pdf} \\
  \includegraphics[width=8cm]{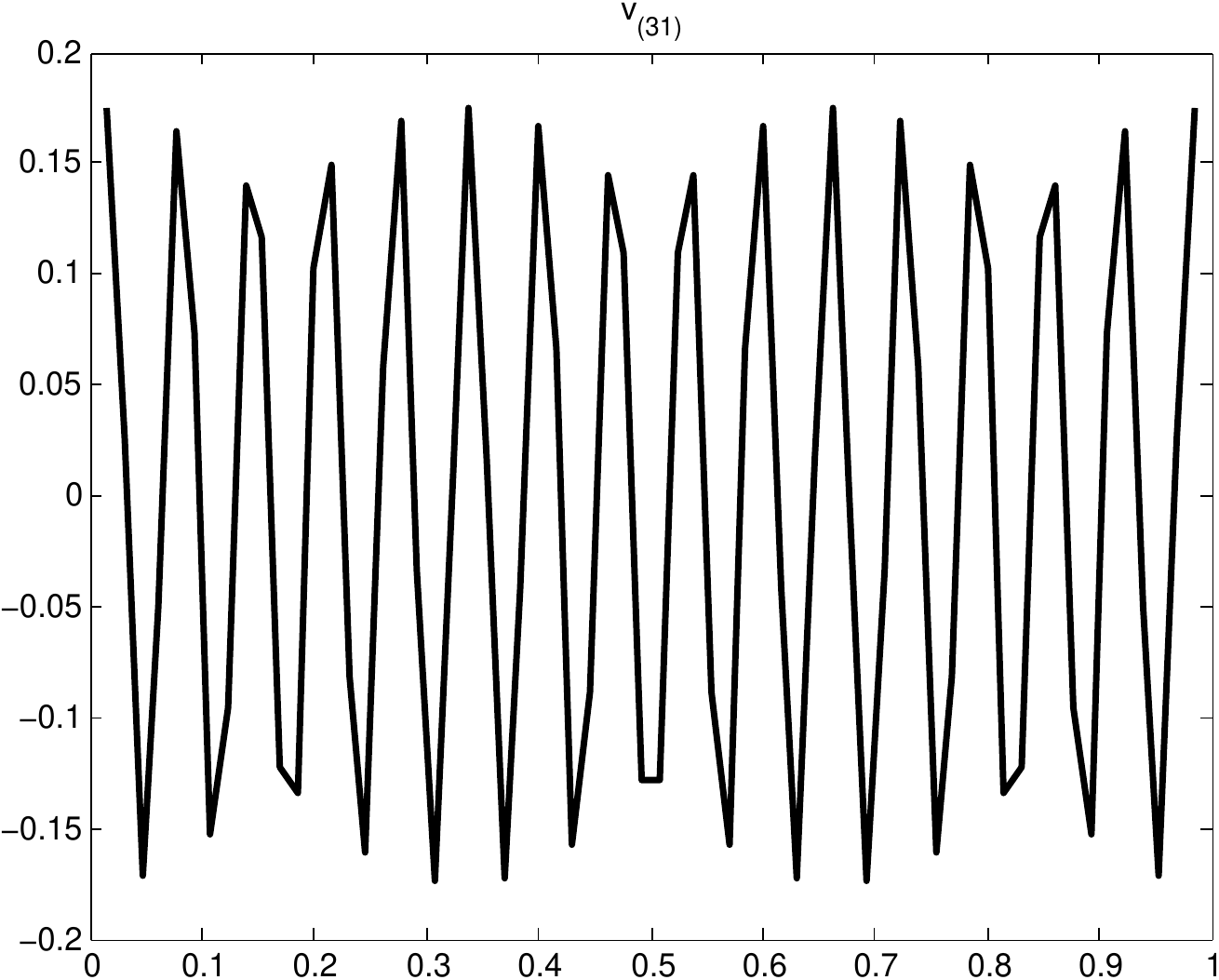} \\%& \includegraphics[width=4cm]{images/v64_64.pdf}
 \end{tabular}
 \caption{
  Eigenfunctions of $A$ when $n=64$.
 }
 \label{fig:eigenfunctions}
\end{figure}

For this problem, it is a simple exercise to show that the eigenfunctions and
eigenvalues of $A$ are
\begin{align}
 v_{(k)i}^h &= \sin (k \pi i h)/h, \; \text{and} \label{eq:exampleEigenfunctions} \\
 \lambda_i^h &= 2 - 2 \cos (\pi i h).
\end{align}
Notice that $i$ directly corresponds to the frequency of the eigenfunctions.
In fact, graph Laplacians always admit a decomposition into a basis of
oscillatory mutually-orthogonal eigenfunctions. This gives us a general notion
of frequency that applies to any graph.
Two of the eigenfunctions of this particular example are depicted in Fig.
\ref{fig:eigenfunctions}, and \textit{exactly} correspond to our usual
interpretation of frequency as Eq. \eqref{eq:exampleEigenfunctions} shows.

The condition number $\kappa(A)$ tells us how difficult it is to accurately
solve a problem involving $A$ (the smaller, the better). Using the small angle approximation, we can see that
$\lambda_1^h \approx \pi^2 h^2$ and $\lambda_n^h \approx 4-h^2$. This gives
a condition number
$\kappa(A) = \lambda_n^h/\lambda_1^h \approx 4/(\pi^2 h^2) = 4(n+1)^2/\pi^2$,
meaning that the condition number is $\order{n^2}$. This implies the difficulty of
solving the problem gets much worse the more variables we have. Notice the cause
is that the lowest-frequency eigenvalue
$\lambda_1^h \rightarrow 0$ as $n \rightarrow \infty$, which makes the condition number
grow unbounded with $n$.
These low-frequency modes are the primary problem we have to grapple if we
wish to solve large graph problems efficiently.

For the Jacobi relaxation, the iteration matrix is
\begin{align}
 R_J = D^{-1}(L+U),
\end{align}
where $D$, $L$ (not to be confused with the Laplacian) and $U$ are the diagonal,
lower-, and upper-triangular parts of $A$. It is a simple exercise to prove that
$\rho(R_J) \approx 1-\pi^2 h^2/2$ since its eigenvalues are
$\mu_i^h=\cos(\pi i h)$, where $i$ again corresponds to the frequency of the
eigenfunction $i$. Recall from \S \ref{subsec:Relaxation}, we want $\rho(R) \ll 1$
for fast convergence. Due to Eq. \eqref{eq:errorIteration}, this means that if
the error $e$ has any low-frequency components, they will converge like
$(1-\pi^2 h^2/2)^j$ (if $j$ is the iteration). This becomes extremely slow
as $n \rightarrow \infty$, $h \rightarrow 0$, and the spectral radius
$\rho(R_J) \rightarrow 1$.

Gauss-Seidel is somewhat better. Its iteration matrix is
\begin{align}
 R_G = (D-L)^{-1}U.
\end{align}
Its eigenvalues are $\mu_i^h = \cos^2(\pi i h)$ for this problem, giving
$\rho(R_G) \leq \rho(R_J)^2$. However, it still has the same basic issue
of slow convergence on low-frequency modes, which gets worse with increasing
problem size $n$.

\begin{figure}
 \centering
 \includegraphics[width=8cm]{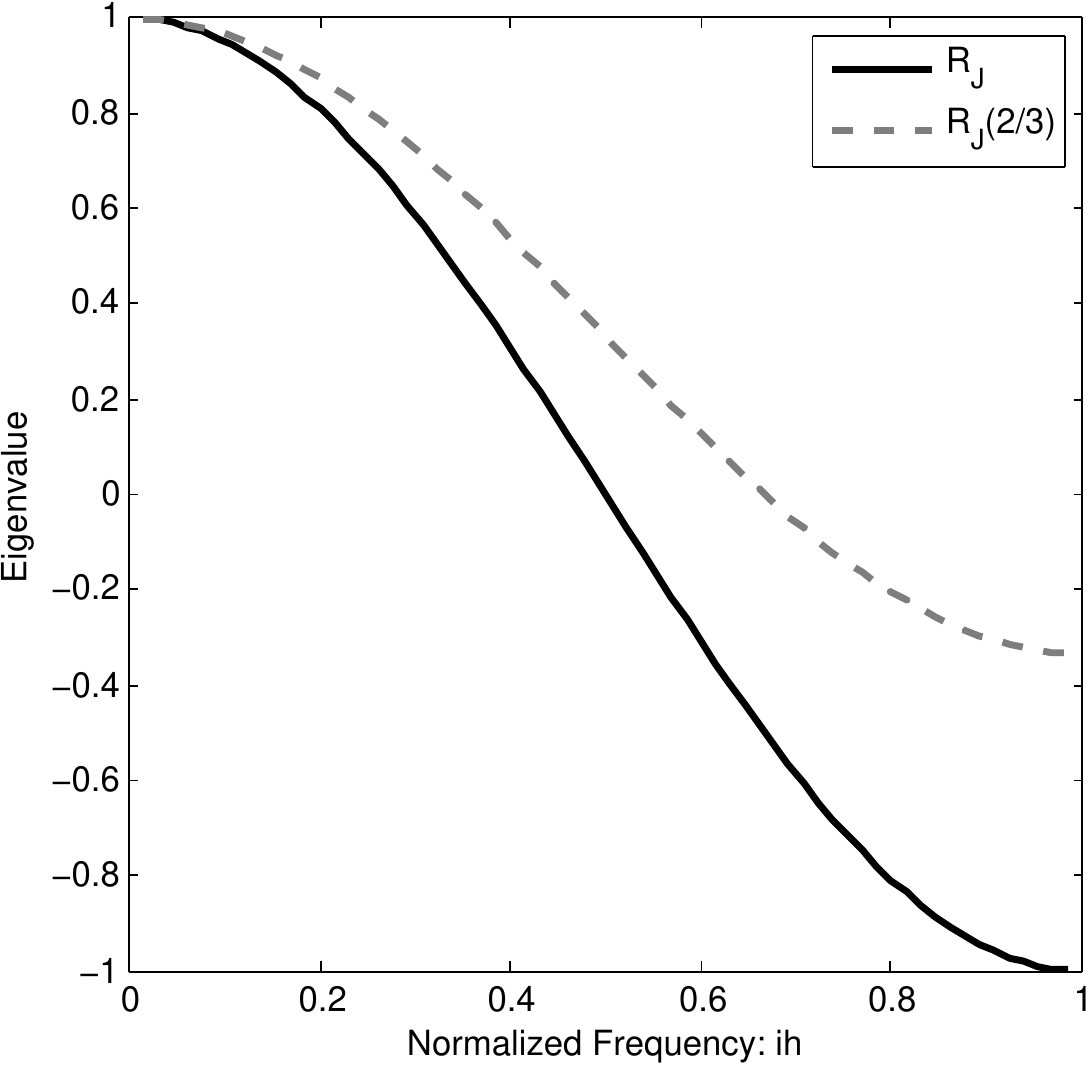}
 \caption{
  Eigenvalues of $R_J$ and $R_J(2/3)$. The latter has high-frequency
  eigenvalues much closer to 0, giving improved high-frequency convergence.
 }
 \label{fig:dampedJacobi}
\end{figure}

As given, both of these methods also have problems with high-frequency modes.
This can be ameliorated with damping:
\begin{align}
 R(\omega) &\defeq (1-\omega)I + \omega R, \; \text{giving} \\
 \mu_i(\omega) &= 1-\omega+\omega \mu_i,
\end{align}
where $\omega$ is the damping parameter, and $\mu_i$ are the eigenvalues of
$R$. By choosing $\omega=2/3$ for the
Jacobi method, the convergence rate for the high-frequency modes significantly
improves as the corresponding eigenvalues decrease in magnitude from near unity
to about $1/3$, as demonstrated in Fig. \ref{fig:dampedJacobi}. A similar
effect occurs with Gauss Seidel.

Gradient descent behaves quite similarly to the Jacobi iteration, and while
conjugate gradient (CG) is a bit harder to analyze in this fashion, it also
displays the same problems in practice. The low-frequency eigenmodes of this
problem are a serious issue if we want to develop a fast solver.

Though we have described a very simple system, we have analagous
problems in matting, just that the constraints are user-supplied and the
Laplacian is usually data-dependent. For matting, most methods have
sidestepped this problem by relying on very dense constraints to resolve most
of the low-frequency components. No one has yet proposed a solver for
Laplacian-based matting which can handle the low-frequency modes efficiently
as the resolution increases and as constraints become sparse.

\subsection{Multigrid Methods}%------------------------------------------------

Multigrid methods are a class of solvers that attempt to fix the problem
of slow low-frequency convergence by integrating a technique called
nested iteration. The basic idea is that, if we can downsample the system,
then the low-frequency modes in high-resolution will become high-frequency
modes in low-resolution, which can be solved with relaxation. A full
treatment is given in \cite{mccormick1987multigrid}.

Without loss of generality, we can assume for simplicity that the image is
square, so that there are $N$ pixels along each axis, $n=N^2$ total pixels,
with $h=1/(N-1)$ being the spacing between the pixels, corresponding to a
continuous image whose domain is $[0,1]^2$. Since we will talk
about different resolutions, we will let $n_h = (1/h+1)^2$ denote the total
number of pixels when the resolution is $h$. We will also assume for
simplicity that the spacings $h$ differ by powers of 2, and therefore the
number of pixels differ by powers of 4. The downsample operators
are linear and denoted by
$I_h^{2h}: \mathbb{R}^{n_h} \rightarrow \mathbb{R}^{n_{2h}}$, and the
corresponding upsample operators are denoted by $I_{2h}^h = (I_h^{2h})^T$.
The choice of these transfer operators is application-specific, and we will
discuss that shortly.

The idea of nested iteration is to consider solutions to
\begin{align}
 I_h^{2h}Au^h = I_h^{2h}f^h,
\end{align}
that is to say a high-resolution solution $u^h$ that solves the downsampled
problem exactly. Such a system is over-determined, so we can restrict our
search to coarse-grid interpolants $u^h = I_{2h}^hu^{2h}$. So, we get
\begin{align}
 I_h^{2h}AI_{2h}^hu^{2h}=I_h^{2h}f^h.
\end{align}
Since $I_{2h}^h = (I_h^{2h})^T$, we can see that
$A^{2h} \defeq I_h^{2h}AI_{2h}^h$ is the equivalent coarse system. Of course
we can repeat this to get $A^{4h}$ and so on. The advantage of a smaller system
is two-fold: first, the system is smaller and therefore is more efficient to
iterate on, and second, the smaller system will converge faster due to better
conditioning.

So, supposing we can exactly solve, say, $u^{16h}$, then we can get an initial
approximation $u_0^h = I_{2h}^hI_{4h}^{2h}\ldots I_{16h}^{8h} u^{16h}$. This
is the simple approach taken by many pyramid schemes, but we can do much
better. The strategy is good if we have no initial guess, but how can this be
useful if we are already given $u_0^h$? The modification is subtle and
important: iterate on the error $e$ instead of the solution variable $u$.
\begin{algorithm}
 \caption{Nested Iteration}
 \label{alg:nested}
 \begin{algorithmic}[1]
  \State $r^h \gets f^h - A^hu^h$
  \State $r^{2h} \gets I_h^{2h} r^h$
  \State Estimate $A^{2h}e^{2h}=r^{2h}$ for $e^{2h}$ using $e_0^{2h}=0$.
  \State $u^h \gets u^h + I_{2h}^he^{2h}$
 \end{algorithmic}
\end{algorithm}

\begin{figure}
 \centering
 \includegraphics[width=8.5cm]{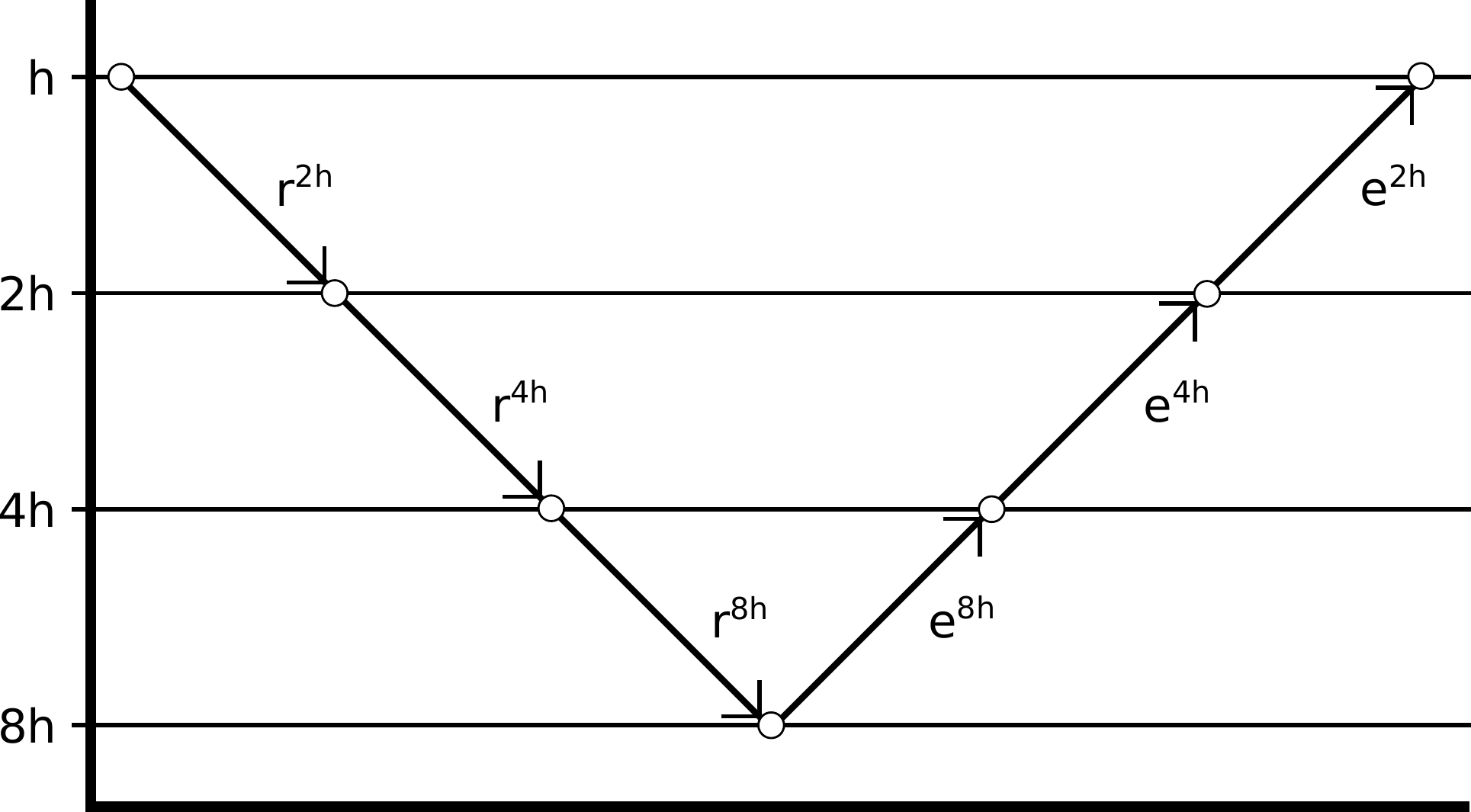}
 \caption{
  Visualization of the v-cycle schedule. The iteration recursively passes down
  the residual to the next lowest resolution and retrieves the error (corresponding
  to lines 6 \& 7 in Alg. \ref{alg:vcycle}). The V shape is the reason for the name.
 }
 \label{fig:vcycleSchedule}
\end{figure}

Of course, step 3 suggests a recursive application of Alg. \ref{alg:nested}.
Notice that if the error is smooth ($e^h \approx I_{2h}^he^{2h}$), then this
works very well since little is lost by solving on the coarse grid. But, if
the error is not smooth, then we make no progress. In this respect,
nested iteration and relaxation are complementary, so it makes a lot of sense
to combine both. This is called \textit{v-cycle}, and is given in Alg.
\ref{alg:vcycle}. A visualization of the algorithm is shown in Fig.
\ref{fig:vcycleSchedule}.

\begin{algorithm}
 \caption{V-Cycle}
 \label{alg:vcycle}
 \begin{algorithmic}[1]
  \Function{vcycle}{$u^h$,$f^h$}
  \If{$h=H$}
   \Return \textit{exact} solution $u^h$
  \EndIf
  \State Relax on $A^h\mathbf{u}^h = f^h$
  \State $r^h \gets f^h - A^hu^h$ (compute residual)
  \State $e^{2h} \gets$ VCYCLE($e^{2h} = 0$,$I_h^{2h}r^h$) (estimate correction)
  \State $u^h \gets u^h + I_{2h}^h e^{2h}$. (apply correction)
  \State Relax on $A^h\mathbf{u}^h = f^h$
  \State \Return $u^h$
  \EndFunction
 \end{algorithmic}
\end{algorithm}

The idea of multigrid is simple: incorporating coarse low-resolution
solutions to the full-resolution scale. The insight is that the low-resolution
parts propagate information quickly, converge quickly, and are cheap to
compute due to reduced system size.

When the problem has some underlying geometry, multigrid methods can
provide superior performance. It turns out that this is not just heuristic,
but also has nice analytic properties. On regular grids, v-cycle is provably
optimal, and converges in a constant number of iterations with respect to the
resolution \cite{gopalakrishnan2008multigrid}.

It is also important to see that the amount of work done per iteration is
$\order{n}$ in the number of variables. To see this, we can take relaxation on the
highest resolution to be 1 unit of work. Looking at Alg. \ref{alg:vcycle}, there
is one relaxation call, followed by recursion, followed by another relaxation call.
In our case, the recursion operates on a problem at $1/4$ size, so we have
the total work as $W(n) = 1 + W(\frac{n}{4}) + 1 = 2 + W(\frac{n}{4})$. In general, the relaxation
costs $n_h/N$, and the expression becomes $W(n_h) = 2\frac{n_h}{n} + W(\frac{n_h}{4})$.
If we continue the expression,
$W(n) = 2 + 2\frac{1}{4} + 2\frac{1}{16} + \cdots = 2\sum_{i=0}^\infty \frac{1}{4}^n$,
which converges to $W(n) = 8/3$. In other words, the work done for v-cycle is
just a constant factor of the work done on a single iteration of the relaxation
at full resolution. Since the relaxation methods chosen are usually $\order{n}$,
then a single iteration of v-cycle is also $\order{n}$. So, we get a lot of
performance without sacrificing complexity.

V-cycle (and the similar w-cycle) multigrid methods work well when the
problem has a regular geometry that allows construction of simple transfer
operators. When the geometry becomes irregular, the transfer operators must
also adapt to preserve the performance benefits. When the geometry is
completely unknown or not present, algebraic multigrid algorithms can be
useful, which attempt to automatically construct problem-specific operators,
giving a kind of ``black box'' solver.

\subsection{Multigrid (Conjugate) Gradient Descent}%---------------------------

Here, we describe a relatively new type of multigrid solver from
\cite{pflaum2008multigrid} that extends multigrid methods to gradient and
conjugate gradient descent. Please note that our notation changes here to
match that of \cite{pflaum2008multigrid} and to generalize the notion of
resolution.

If the solution space has a notion of resolution on $\ell$ total levels, with downsample operators
$I_\ell^i: \mathbb{R}^{n_\ell} \rightarrow \mathbb{R}^{n_i}$, and upsample operators
$I_i^\ell: \mathbb{R}^{n_i} \rightarrow \mathbb{R}^{n_\ell}$
($n_i < n_\ell \; \forall i < \ell$), then define the multi-level residuum space
on $x^k$ at the $k^\text{th}$ iteration as
\begin{align}
 \mathcal{R}^k &= \spanop \left\{ r_\ell^k, \dotsc, r_1^k \right\}, \; \text{where}\\
 r_i^k &= I_i^\ell I_\ell^i r_\ell^k, \; \forall i < \ell \; \text{and}\\
 r_\ell^k &= f - Au^k,
\end{align}
where $r_\ell^k$ is the residual at the finest level at iteration $k$,
and $r_i^k$ is the projection of the fine residual to the coarser levels. This
residuum space changes at each iteration $k$.
Gradient descent in this multi-level residuum space would be to find correction
direction
$d \in \mathcal{R}^k$ s.t. $\langle u^k+d, q \rangle_A = f^Tq \; \forall q \in
\mathcal{R}^k$. In other words, the correction direction is a linear combination
of low-resolution residual vectors at iteration $k$, forcing the low-resolution
components of the solution to appear quickly.

To find step direction $d$ efficiently, one needs to orthogonalize $\mathcal{R}^k$ by the
Gram-Schmidt procedure:
\begin{align}
 \hat{d}_i &= r_i^k - \sum_{j=1}^{i-1} \inner{r_i^k}{d_j}_A d_j, \; \forall i \in 1,\dotsc,\ell \\
 d_i &= \frac{\hat{d}_i}{\norm{\hat{d}_i}_A}, \\
 d &= \sum_{i=1}^\ell \inner{r_\ell^k}{d_i} d_i.
\end{align}
These equations give an $\order{n_\ell \times \ell}$ algorithm if implemented na\"{\i}vely.
But, since all of the inner products can be performed at their respective
downsampled resolutions, the efficiency can be improved to $\order{ n_\ell }$.
%The full procedure is given in Alg. \ref{alg:mgGradient} from \cite{pflaum2008multigrid}.
To get an idea how this can be done, consider
computing the inner product $\inner{d_i}{d_i}_A$. First see
that \cite{pflaum2008multigrid} guarantees the existence of the low-resolution
correction direction vector
$d_i^{\text{co}} \in \mathbb{R}^{n_i}$ such that it can reconstruct the
full-resolution direction by upsampling, i.e. $d_i = I_i^\ell d_i^{\text{co}}$.
Now, notice that
\begin{align}
 \inner{d_i}{d_i}_A &= d_i^T A d_i \nonumber \\
   &= (I_i^\ell d_i^{\text{co}})^T A (I_i^\ell d_i^{\text{co}}) \\
   &= (d_i^{\text{co}})^T (I_\ell^i A I_i^\ell) d_i^{\text{co}} \\
   & = (d_i^{\text{co}})^T A_i d_i^{\text{co}},
\end{align}
Where $A_i$ is a lower-resolution system. Since $A_i$ is only $n_i \times n_i$, and $n_i \ll n_\ell$ if $i < \ell$,
then working directly on the downsampled space allows us to perform such
inner products much more efficiently.

Standard gradient descent is simple to implement, but it tends to display poor
convergence properties. This is because there is no relationship between
the correction spaces in each iteration, which may be largely
redundant, causing unnecessary work. This problem is fixed with
conjugate gradient (CG) descent.

In standard conjugate gradient descent, the idea is that the correction
direction for iteration $k$ is $A$-orthogonal (conjugate) to the previous correction
directions, i.e.
\begin{align}
 \mathcal{D}^k \perp_A \mathcal{D}^{k-1},
\end{align}
where $\mathcal{D}^k$ is iteration $k$'s correction space analagous to
$\mathcal{R}^k$. This leads to faster convergence than gradient descent, due to
never performing corrections that interfere with each other.

The conjugate gradient approach can also be adapted to the residuum
space. It would require that $d \in \mathcal{D}^k$, where
$\spanop\{\mathcal{D}^k,\mathcal{D}^{k-1}\} =
\spanop\{\mathcal{R}^k,\mathcal{D}^{k-1}\}$ and
$\mathcal{D}^k \perp_A \mathcal{D}^{k-1}$. However, as shown by
\cite{pflaum2008multigrid}, the resulting orthogonalization procedure would
cause the algorithm complexity to be $\order{n_\ell \times \ell}$, even when performing
the multiplications in the downsampled resolution spaces. By slightly relaxing
the orthogonality condition, \cite{pflaum2008multigrid} develops a multigrid
conjugate gradient (MGCG) algorithm
with complexity $\order{n_\ell}$.
\begin{align}
 \mathcal{D}^k &= \spanop \left\{ d_1^k, \dotsc, d_l^k  \right\} \\
 \norm{d_i^k}_A &= 1 \; \forall i \in 1,\dotsc,\ell \\
 d_i^k &\perp_A d_j^k \; \forall i \neq j \\
 \mathcal{D}^1 &= \mathcal{R}^1 \\
 d_i^{k-1} &\perp_A d_j^{k} \; \forall k > 1, i \leq j \\
 \mathcal{R}^k + \mathcal{D}^{k-1} &= \mathcal{D}^k + \mathcal{D}^{k-1}
\end{align}
We present Alg. \ref{alg:mgCG} in the appendix, which fixes some algorithmic
errors made in \cite{pflaum2008multigrid}.

In multigrid algorithms such as these, we have freedom to construct the
transfer operators. If we want to take full advantage of such algorithms, we
must take into account our prior knowledge of the solution when designing the
operators.

\section{Evaluated Solution Methods}%==========================================

\subsection{Conjugate Gradient}%-----------------------------------------------

CG and variants thereof are commonly used to solve the matting problem, so we
examine the properties of the classic CG method alongside the multigrid
methods.

\subsection{V-cycle}%----------------------------------------------------------

In order to construct a v-cycle algorithm, we need to choose a relaxation
method and construct transfer operators. We chose Gauss-Seidel as the
relaxation method, since it is both simple and efficient. Through experiment,
we found that simple bilinear transfer operators (also called
\textit{full-weighting}) work very well for the matting problem. The downsample
operator $I_h^{2h}$ is represented by the stencil
\begin{align}
 \begin{bmatrix}
  \frac{1}{16} & \frac{1}{8} & \frac{1}{16} \\
  \frac{1}{8} & \frac{1}{4} & \frac{1}{8} \\
  \frac{1}{16} & \frac{1}{8} & \frac{1}{16}
 \end{bmatrix}. \nonumber
\end{align}
If the stencil is centered on a fine grid point, then the numbers represent the
contribution of the 9 fine grid points to their corresponding coarse grid point.
The corresponding upsample operator $I_{2h}^h$ is represented by the stencil
\begin{align}
 \left]
 \begin{matrix}
  \frac{1}{4} & \frac{1}{2} & \frac{1}{4} \\
  \frac{1}{2} & 1 & \frac{1}{2} \\
  \frac{1}{4} & \frac{1}{2} & \frac{1}{4}
 \end{matrix}
 \right[. \nonumber
\end{align}
This stencil is imagined to be centered on a fine grid point, and its numbers
represent the contribution of the corresponding coarse grid value to the 9
fine grid points. The algorithm is then given by Alg. \ref{alg:vcycle}.

\subsection{Multigrid Conjugate Gradient}%-------------------------------------

We also examine the multigrid conjugate gradient algorithm of
\cite{pflaum2008multigrid}, which is a more recent multigrid variant, and
seems like a natural extension of CG. For this algorithm, we find that the bilinear
(full-weighting) transfer operators also work very well. The algorithm
is given by Alg. \ref{alg:mgCG}, where we give the full pseudo-code as the
original pseudo-code given in \cite{pflaum2008multigrid} is erroneous.

\section{Experiments}%=========================================================

\begin{figure}
 \centering
 \begin{tabular}{c}
  \includegraphics[width=8.5cm]{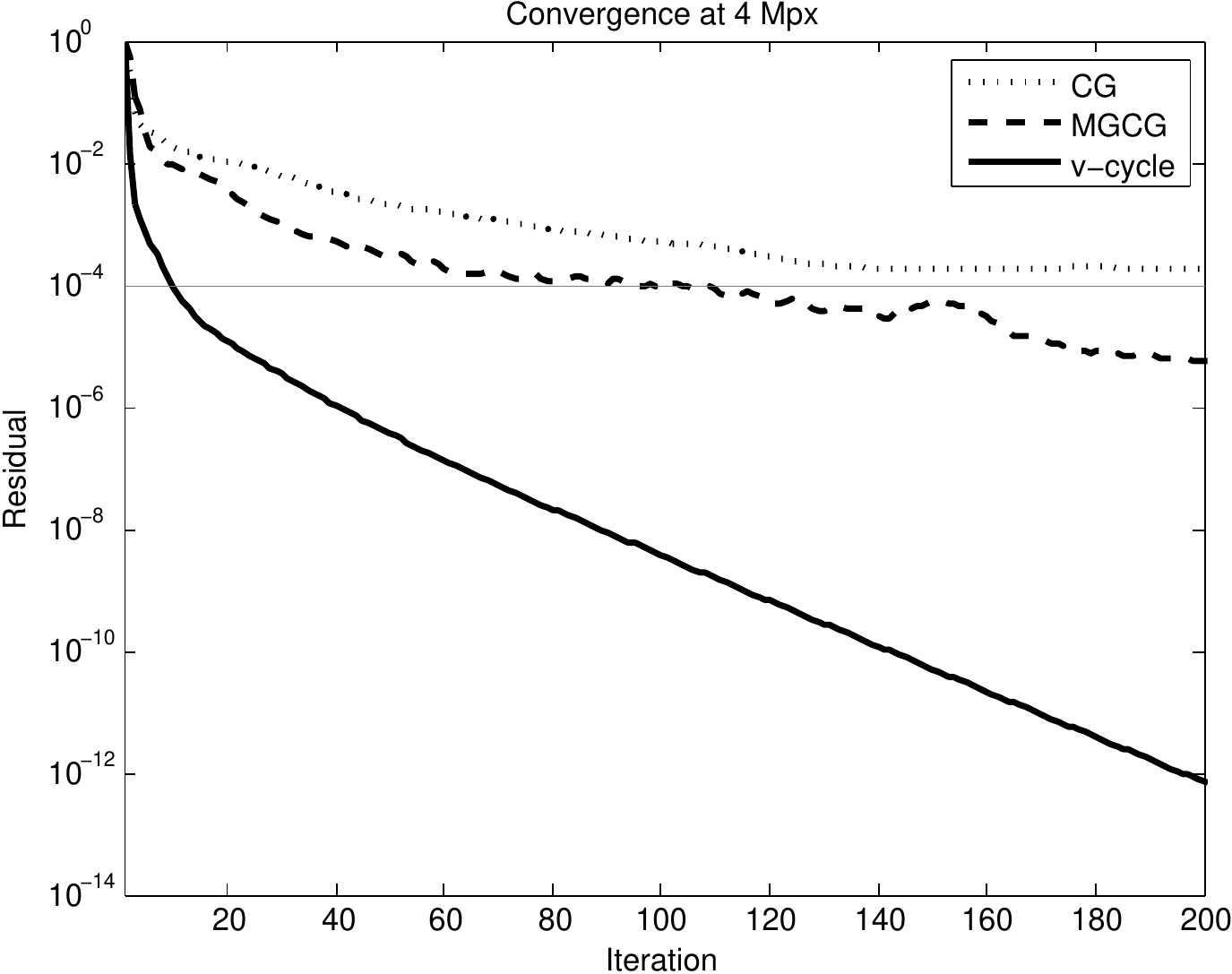}
 \end{tabular}
 \caption{
  Convergence on a 4 Mpx image. Horizontal line denotes proposed termination
  value of $10^{-4}$.
 }
 \label{fig:res4Mpx}
\end{figure}

\begin{figure}
 \centering
 \begin{tabular}{c}
  \includegraphics[width=8cm]{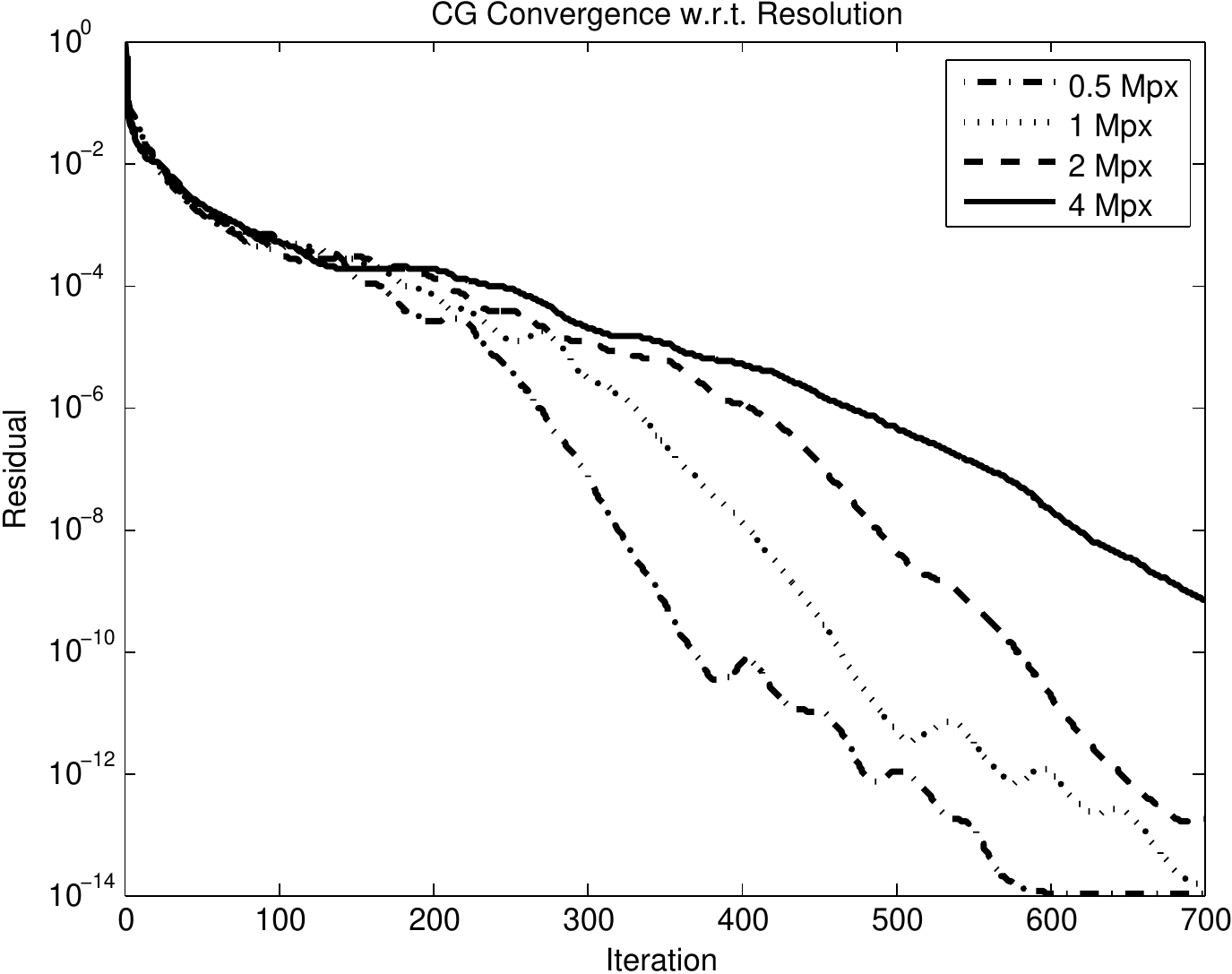} \\
  (a) \\
  \includegraphics[width=8cm]{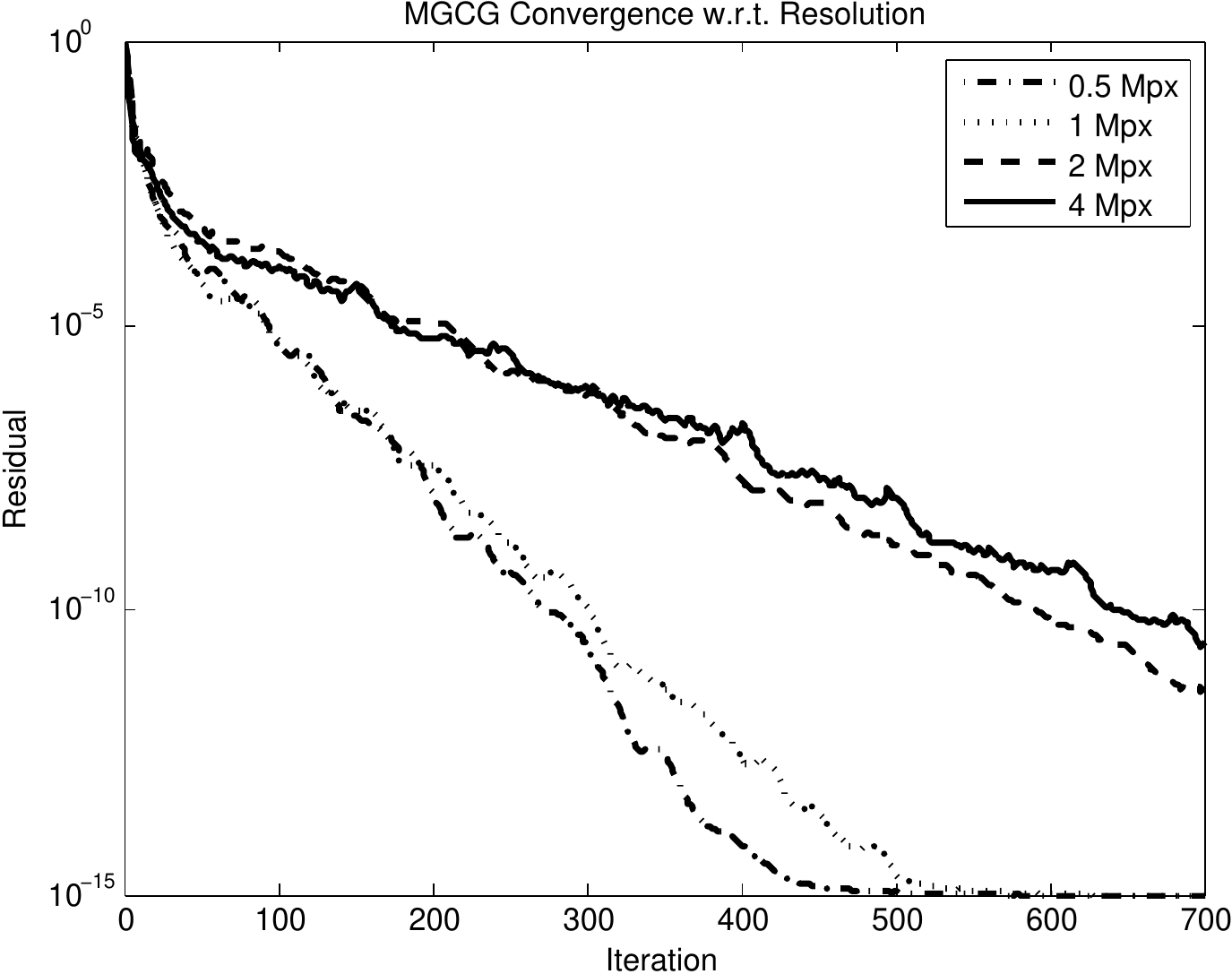} \\
  (b) \\
  \includegraphics[width=8cm]{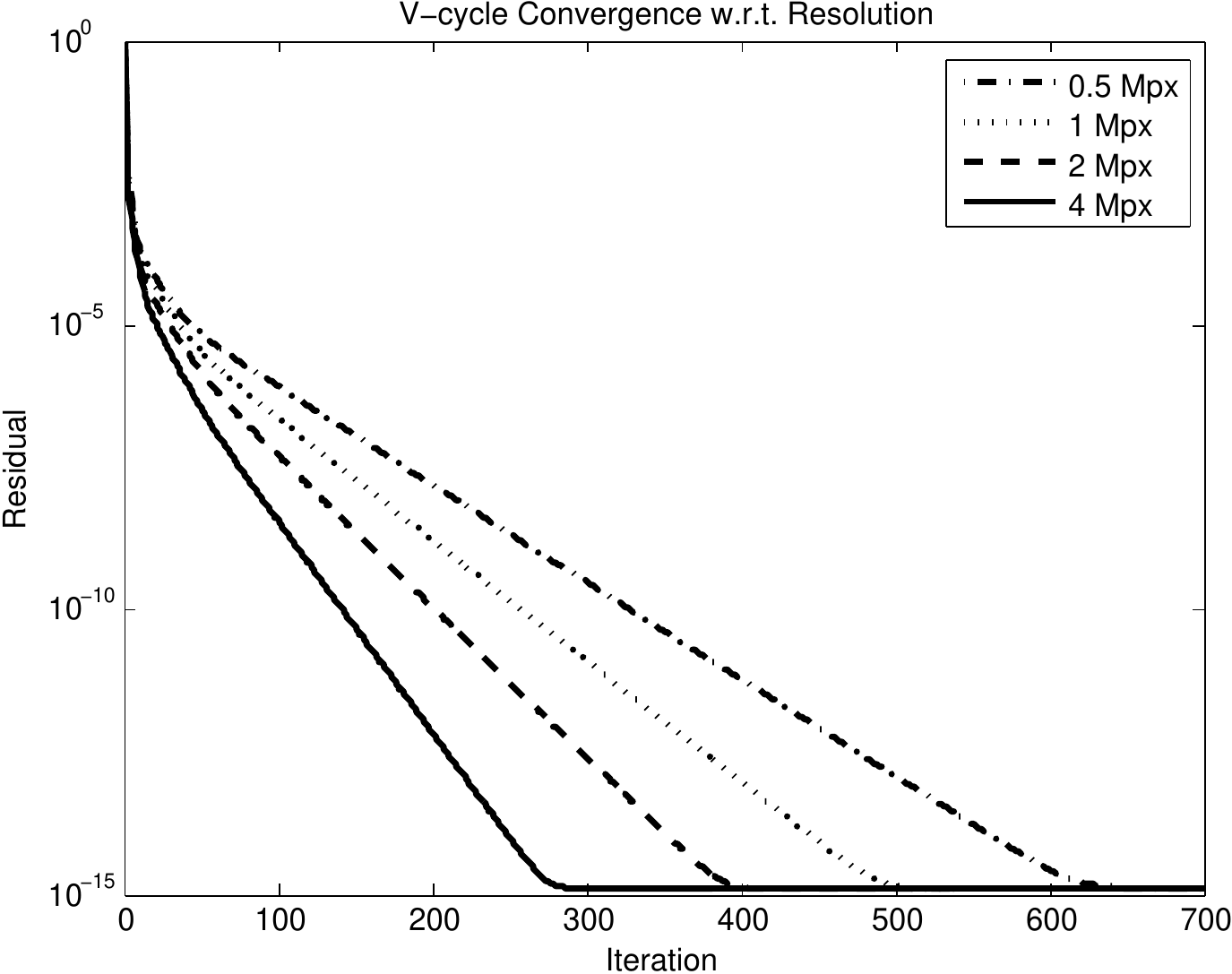} \\
  (c)
 \end{tabular}
 \caption{
  CG (a) \& MGCG (b) slow down as resolution increases, while v-cycle (c)
  \textit{speeds up}.
 }
 \label{fig:resVcycle}
\end{figure}

\begin{table}
 \centering
 \begin{tabular}{|r|r|r|r|}
  \hline
  Resolution & CG & MGCG & v-cycle \\
  \hline
  0.5 Mpx & 0.594 & 0.628 & 0.019 \\
  1 Mpx   & 0.357 & 0.430 & 0.017 \\
  2 Mpx   & 0.325 & 0.725 & 0.017 \\
  4 Mpx   & 0.314 & 0.385 & 0.015 \\
  \hline
 \end{tabular}
 \caption{
  Initial convergence rates $\rho_0$ on image 5 in
  \cite{rhemann2009perceptually}. Lower is better.
 }
 \label{tab:convRates}
\end{table}

\begin{table*}
 \centering
 \begin{tabular}{|r|rrr|rrr|rrr|}
  \hline
        & \multicolumn{3}{c}{CG} & \multicolumn{3}{|c|}{MGCG} & \multicolumn{3}{c|}{v-cycle} \\
  Image & 1 Mpx & 2 Mpx & 4 Mpx & 1 Mpx & 2 Mpx & 4 Mpx & 1 Mpx & 2 Mpx & 4 Mpx \\
  \hline
  1  & 143 & 159 & 179 &  81 &  81 &  94 & 12 & 11 & 11 \\
  2  & 151 & 176 & 185 &  73 &  86 &  99 & 12 & 10 & 10 \\
  3  & 215 & 280 & 334 & 103 & 108 & 118 & 12 &  9 &  8 \\
  4  & 345 & 439 & 542 & 143 & 171 & 230 & 16 & 14 & 11 \\
  5  & 180 & 211 & 236 &  44 & 120 &  95 & 13 & 12 & 10 \\
  6  & 126 & 148 & 180 &  65 &  91 &  61 & 11 & 10 &  9 \\
  7  & 152 & 179 & 215 &  86 &  84 &  96 & 11 & 10 &  8 \\
  8  & 314 & 385 & 468 & 135 & 167 & 180 & 16 & 12 &  9 \\
  9  & 237 & 285 & 315 &  87 & 119 &  91 & 12 & 11 &  8 \\
  10 & 122 & 149 & 180 &  72 &  72 &  95 & 10 &  9 &  8 \\
  11 & 171 & 192 &  78 &  77 &  96 &  32 & 12 & 11 &  7 \\
  12 & 127 & 151 & 179 &  65 &  86 &  77 &  9 &  8 &  7 \\
  13 & 266 & 291 & 317 & 143 & 156 & 159 & 18 & 15 & 11 \\
  14 & 125 & 145 & 175 &  76 &  79 &  83 & 13 & 12 & 11 \\
  15 & 129 & 156 & 190 &  67 &  69 & 105 & 12 & 12 & 11 \\
  16 & 292 & 374 & 468 & 123 & 171 & 178 & 15 & 14 & 12 \\
  17 & 153 & 190 & 230 &  63 &  81 &  86 & 10 &  8 &  7 \\
  18 & 183 & 238 & 289 &  86 & 104 & 107 & 26 & 22 & 16 \\
  19 &  91 & 104 & 116 &  52 &  58 &  61 & 12 & 10 &  8 \\
  20 & 137 & 164 & 199 &  81 &  82 & 101 & 12 &  9 &  7 \\
  21 & 195 & 288 & 263 & 118 &  83 & 137 & 22 & 16 & 15 \\
  22 & 118 & 139 & 161 &  62 &  75 &  72 & 11 &  9 &  7 \\
  23 & 140 & 166 & 196 &  64 &  77 &  74 & 11 &  9 &  8 \\
  24 & 213 & 266 & 325 &  76 & 131 & 104 & 14 & 14 & 10 \\
  25 & 319 & 455 & 629 & 142 & 170 & 249 & 39 & 31 & 24 \\
  26 & 370 & 432 & 478 & 167 & 192 & 229 & 41 & 32 & 24 \\
  27 & 271 & 304 & 349 & 121 & 144 & 147 & 17 & 15 & 12 \\
  \hline
 \end{tabular}
 \caption{
  Iterations to convergence (residual less than $10^{-4}$) at 1, 2, and 4 Mpx
  on images from \cite{rhemann2009perceptually}. CG and MGCG require more
  iterations as resolution increases while v-cycle requires less.
 }
 \label{tab:allImages}
\end{table*}

\begin{table}
 \centering
 \begin{tabular}{|r|r|r|}
  \hline
  Image & $a$ & $p$ \\
  \hline
   1 & 11.8 & -0.065 \\
   2 & 11.7 & -0.140 \\
   3 & 11.8 & -0.309 \\
   4 & 16.2 & -0.261 \\
   5 & 13.2 & -0.183 \\
   6 & 11.0 & -0.144 \\
   7 & 11.2 & -0.220 \\
   8 & 16.0 & -0.415 \\
   9 & 12.3 & -0.270 \\
  10 & 10.0 & -0.160 \\
  11 & 12.5 & -0.343 \\
  12 & 9.02 & -0.180 \\
  13 & 18.2 & -0.340 \\
  14 & 13.0 & -0.120 \\
  15 & 12.2 & -0.061 \\
  16 & 15.2 & -0.156 \\
  17 & 9.90 & -0.265 \\
  18 & 26.4 & -0.333 \\
  19 & 12.1 & -0.289 \\
  20 & 12.0 & -0.394 \\
  21 & 21.4 & -0.300 \\
  22 & 11.1 & -0.321 \\
  23 & 10.9 & -0.236 \\
  24 & 14.6 & -0.216 \\
  25 & 39.1 & -0.347 \\
  26 & 41.2 & -0.381 \\
  27 & 17.2 & -0.243 \\
  \hline
 \end{tabular}
 \caption{
  Fitting of v-cycle required iterations to $an^p$ (power law), with $n$ being problem size
  in Mpx. Average $p$ is $E[p] = -0.248$, meaning this solver is sublinear
  in $n$.
 }
 \label{tab:powerFit}
\end{table}

In order to compare the solution methods, we choose the matting Laplacian
proposed by \cite{levin2008closed}, as it is popular and provides source code.
Please note that we are comparing the solvers, and not the performance
of the matting Laplacian itself, as these solvers would equally apply to other
matting Laplacians. We set their $\epsilon$ parameter to $10^{-3}$, and we form
our linear system according to Eq. \eqref{eq:linearSystem} with $\gamma=1$.

All methods are given the initial guess $\alpha_0=0$. To make the objective
value comparable across methods and resolutions, we take
$\norm{f-A\alpha_i}_2/\norm{f-A\alpha_0}_2 = \norm{f-A\alpha_i}_2/\norm{f}_2$
as the normalized residual at iteration $i$, or \textit{residual} for short.
We set the termination condition for all methods at all resolutions to be the
descent of the residual below $10^{-4}$.

First, we evaluate the convergence of conjugate gradient, multigrid conjugate
gradient, and v-cycle. Typical behavior on the dataset
\cite{rhemann2009perceptually} is shown in Fig. \ref{fig:res4Mpx} at 4 Mpx
resolution. CG takes many iterations to converge, MGCG performs somewhat
better, and v-cycle performs best.

Next, we demonstrate in Fig. \ref{fig:resVcycle} that the number of iterations v-cycle requires to
converge on the matting problem does not increase with the resolution. In fact,
we discover that v-cycle requires \textit{fewer} iterations as the resolution
increases. This supports a sub-linear complexity of v-cycle on this problem in
practice since each iteration is $\order{n}$.

We list a table of the convergence behavior of the algorithms on all
images provided in the dataset in Table \ref{tab:allImages}. This demonstrates
that MGCG performs decently better than CG, and that v-cycle always requires
few iterations as the resolution increases. By fitting a power law to the
required iterations in Table \ref{tab:powerFit}, we find the average number of
required v-cycle iterations is $\order{n^{-0.248}}$. Since each iteration is
$\order{n}$, this means that this is the first solver demonstrated to have
sublinear time complexity on the matting problem. Specifically, the time
complexity is $\order{n^{0.752}}$ on average. Theoretical guarantees
\cite{gopalakrishnan2008multigrid} ensure that the worst case time complexity
is $\order{n}$.

Though surprising, the sub-linear behavior is quite understandable. Most objects to be matted
are opaque with a boundary isomorphic to a line segment. The interior and
exterior of the object have completely constant $\alpha$ locally, which is
very easy for v-cycle to handle at its lower levels, since that is where
the low frequency portions of the solution get solved. This leaves the boundary
pixels get handled at the highest resolutions, since edges are high-frequency.
These edge pixels contribute to the residual value, but
since the ratio of boundary to non-boundary pixels is $\order{n^{-0.5}}$,
tending to 0 with large $n$, the residual tends to decrease more rapidly per
iteration as the resolution increases as proportiontely more of the problem
gets solved in lower resolution.

Another way to view it is that there is a fixed amount of information present
in the underlying image, and increasing the number of pixels past some point does not really
add any additional information. V-cycle is simply the first algorithm studied to
exploit this fact by focusing its effort preferentially on the low resolution
parts of the solution.

Perhaps if we had enough computational resources and a high-enough resolution,
v-cycle could drop the residual to machine epsilon in a single iteration. At
that point, we would again have a linear complexity solver,
as we would always have to do 1 iteration at $\order{n}$ cost. In any case, $\order{n}$
is \textit{worst} case behavior as guaranteed by \cite{gopalakrishnan2008multigrid},
which is still much better than any solver proposed to date like
\cite{he2010fast} and \cite{xiao2014fast} (which are at best $\order{n^2}$).
We would also like to mention that the per-iteration work done by the proposed
method is constant with respect to the number of unknowns, unlike
most solvers including \cite{he2010fast} and \cite{xiao2014fast}.

Table \ref{tab:convRates} lists the initial convergence rates $\rho_0$ (the
factor by which the residual is reduced per iteration).
\cite{gopalakrishnan2008multigrid} guarantees that the convergence rate for
v-cycle is at most $\order{1}$ and bounded away from 1. We discover that on
the matting problem the convergence rate is \textit{sub}-constant. Yet again,
v-cycle dramatically overshadows other solvers.

Although a MATLAB implementation of \cite{levin2008closed} is available, we
found it to use much more memory than we believe it should. We therefore developed
our own implementation in C++ using efficient algorithms and data structures
to run experiments at increasing resolution. Since there are 25 bands in the
Laplacian, and 4 bytes per floating point entry, the storage required for the
matrix in \eqref{eq:linearSystem} is 95 MB/Mpx, and the total storage including the downsampled matrices
is about 127 MB/Mpx. The time to generate the Laplacian is 0.8 s/Mpx. The CPU
implementation runs at 46 Gauss-Seidel iterations/s/Mpx
and 11 v-cycle iterations/s/Mpx. Assuming 10 iterations of v-cycle are run,
which is reasonable from Table \ref{tab:allImages}, our solver runs in
about 1.1 s/Mpx, which is nearly 5 times faster than \cite{he2010fast} at low
resolutions, and even faster at higher resolutions. Upon acceptance of this
paper, we will post all the code under an open-source license.

\begin{table}
 \centering
 \begin{tabular}{|r|rrrr|}
  \hline
  Method & SAD & MSE & Gradient & Connectivity \\
  \hline
  Ours                   & \textbf{11.1} & \textbf{10.8} & \textbf{12.9} & 9.1 \\
  \cite{levin2008closed} & 14 & 13.9 & 14.5 & \textbf{6.4} \\
  \cite{he2010fast}      & 17 & 15.8 & 16.2 & 9.2 \\
  \hline
 \end{tabular}
 \caption{ Average rank in benchmark \cite{rhemann2009perceptually} on 11 Mar 2013
           with respect to different error metrics.}
 \label{tab:benchmark}
\end{table}

Although we consider this to be a matting solver rather than a new matting
algorithm, we still submitted it to the online benchmark
\cite{rhemann2009perceptually} using \cite{levin2008closed}'s Laplacian.
The results presented in Table \ref{tab:benchmark} show some improvement over
the base method \cite{levin2008closed}, and more improvement over another fast
method \cite{he2010fast}. In theory, we should reach the exact same solution as
\cite{levin2008closed} in this experiment, so we believe any performance gain
is due to a different choice of their $\epsilon$ parameter.

\section{Conclusion}%==========================================================

We have thoroughly analyzed the performance of two multigrid algorithms,
discovering that one exhibits sub-linear complexity on the matting
problem. The $\order{n^{0.752}}$ average time complexity is by far the most
efficient demonstrated to date. Further benefits of this approach
include the ability to directly substitute any matting Laplacian of choice,
and the capability to solve the original full-scale problem exactly without
heuristics.

\ifCLASSOPTIONcompsoc
  \section*{Acknowledgments}
\else
  \section*{Acknowledgment}
\fi

This work was supported in part by National Science Foundation grant
IIS-0347877, IIS-0916607, and US Army Research Laboratory and the US Army
Research Office under grant ARO W911NF-08-1-0504.

\appendices%===================================================================

\section{Multigrid Algorithms}%------------------------------------------------
\begin{algorithm}
\caption{Multigrid gradient descent.}
\label{alg:mgGradient}
\begin{algorithmic}[1]
 \While { $numIter > 0$ }
   \State $numIter \gets numIter - 1$
   \State $r_\ell \gets b - Ax$
   \State $d_\ell \gets r_\ell$
   \For{ $i = \ell-1..1$ }
     \State $r_i \gets I_{i+1}^i r_{i+1}$
     \State $d_i \gets r_i$
   \EndFor
   
   \For{ $i = 1..\ell$ }
     \State $k_i \gets A_i d_i$
     \For{ $j = i..2$ }
       \State $k_{j-1} \gets I_j^{j-1} k_j$
     \EndFor
     \State $s_1 = 0$
     \For{ $j = 1..i-1$ }
       \State $s_j \gets s_j + (k_j^Td_j) d_j $
       \State $s_{j+1} \gets I_j^{j+1} s_j$
     \EndFor
     \State $d_i \gets d_i - s_i$
     \State $k_i \gets A_i d_i$
     \State $d_i \gets d_i (k_i^Td_i)^{-\frac{1}{2}}$
   \EndFor
   
   \State $s_1 \gets (d_1^Tr_1) d_1$
   \For{ $i=2..\ell$ }
     \State $s_i \gets I_{i-1}^i s_{i-1}$
     \State $s_i \gets s_i + d_i$
   \EndFor
   
   \State $x \gets x + s_\ell$
 \EndWhile
\end{algorithmic}
\end{algorithm}

\begin{algorithm}
\caption{Multigrid CG descent.}
\label{alg:mgCG}
\begin{algorithmic}[1]
 \State {\small Do one iteration of multigrid gradient descent.}
 \While { $numIter > 0$ }
   \State $numIter \gets numIter - 1$
   \State $r_\ell \gets b - Ax$
   \State $d_\ell^\text{new} \gets r_\ell$
   \For{ $i = \ell-1..1$ }
     \State $r_i \gets I_{i+1}^i r_{i+1}$
     \State $d_i^\text{new} \gets r_i$
   \EndFor
   
   \For{ $i = 1..\ell$ }
     \State $k_i \gets A_i d_i$
     \For{ $j = i..2$ }
       \State $k_{j-1} \gets I_j^{j-1} k_j$
     \EndFor
     \State $s_1 = 0$
     \For{ $j = 1..i-1$ }
       \State $s_j \gets s_j + (k_j^Td_j^\text{new}) d_j^\text{new} $
       \State $s_{j+1} \gets I_j^{j+1} s_j$
     \EndFor
     \State $d_i \gets d_i - s_i$
     \State $k_i \gets A_i d_i$
     \State $d_i \gets d_i (k_i^Td_i)^{-\frac{1}{2}}$
     
     \State $k_i \gets A_i d_i^\text{new}$
     \For{ $j = i..2$ }
       \State $k_{j-1} \gets I_j^{j-1} k_j$
     \EndFor
     \State $s_1 \gets 0$
     \For{ $j = 1..i-1$ }
       \State $s_j \gets s_j + d_j^\text{new}(k_j^Td_j^\text{new})$
       \State $s_j \gets s_j + d_j (k_j^Td_j^\text{new})$
       \State $s_{j+1} \gets I_j^{j+1} s_j$
     \EndFor

     \State $d_i^\text{new} \gets d_i^\text{new} - s_i$
     \State $s_i \gets d_i^\text{new} - d_i (k_i^Td_i)$
     
     \If{ $\norm{s_i}_\infty > \epsilon \norm{d_i^\text{new}}_\infty$ }
       \State $d_i^\text{new} \gets s_i$
     \Else
       \State $d_i^\text{new} \gets d_i$
       \State $d_i \gets 0$
     \EndIf
     
     \State $k_i \gets A_i d_i^\text{new}$
     \State $d_i^\text{new} \gets d_i^\text{new} (k_i^Td_i^\text{new})^{-1/2}$
   \EndFor
     
   \State $d_1 \gets d_1^\text{new}$
   \State $s_1 \gets d_1 (d_1^Tr_1)$
   
% This splits up the algorithm so that it doesn't overfill a single page.
% \algstore{mgCG}
% \end{algorithmic}
% \end{algorithm}
% \begin{algorithm}
% \begin{algorithmic}[1]
% \algrestore{mgCG}
   
   \For{ $i = 2..\ell$ }
     \State $d_i \gets d_i^\text{new}$
     \State $s_i \gets I_{i-1}^i s_{i-1}$
     \State $s_i \gets s_i + d_i (d_i^Tr_i)$
   \EndFor
     
   \State $x \gets x + s_\ell$
 \EndWhile
\end{algorithmic}
\end{algorithm}

%------------------------------------------------------------------------------

\bibliographystyle{IEEEtran}
\bibliography{IEEEabrv,MultiGridMatting}

\begin{thebibliography}{10}
\providecommand{\url}[1]{#1}
\csname url@samestyle\endcsname
\providecommand{\newblock}{\relax}
\providecommand{\bibinfo}[2]{#2}
\providecommand{\BIBentrySTDinterwordspacing}{\spaceskip=0pt\relax}
\providecommand{\BIBentryALTinterwordstretchfactor}{4}
\providecommand{\BIBentryALTinterwordspacing}{\spaceskip=\fontdimen2\font plus
\BIBentryALTinterwordstretchfactor\fontdimen3\font minus
  \fontdimen4\font\relax}
\providecommand{\BIBforeignlanguage}[2]{{%
\expandafter\ifx\csname l@#1\endcsname\relax
\typeout{** WARNING: IEEEtran.bst: No hyphenation pattern has been}%
\typeout{** loaded for the language `#1'. Using the pattern for}%
\typeout{** the default language instead.}%
\else
\language=\csname l@#1\endcsname
\fi
#2}}
\providecommand{\BIBdecl}{\relax}
\BIBdecl

\bibitem{he2009single}
K.~He, J.~Sun, and X.~Tang, ``Single image haze removal using dark channel
  prior,'' in \emph{Computer Vision and Pattern Recognition, 2009. CVPR 2009.
  IEEE Conference on}.\hskip 1em plus 0.5em minus 0.4em\relax IEEE, 2009, pp.
  1956--1963.

\bibitem{dai2008motion}
S.~Dai and Y.~Wu, ``Motion from blur,'' in \emph{Computer Vision and Pattern
  Recognition, 2008. CVPR 2008. IEEE Conference on}.\hskip 1em plus 0.5em minus
  0.4em\relax IEEE, 2008, pp. 1--8.

\bibitem{fan2010closed}
J.~Fan, X.~Shen, and Y.~Wu, ``Closed-loop adaptation for robust tracking,'' in
  \emph{Computer Vision--ECCV 2010}.\hskip 1em plus 0.5em minus 0.4em\relax
  Springer, 2010, pp. 411--424.

\bibitem{levin2008closed}
A.~Levin, D.~Lischinski, and Y.~Weiss, ``{A closed-form solution to natural
  image matting},'' \emph{IEEE Transactions on Pattern Analysis and Machine
  Intelligence}, vol.~30, no.~2, pp. 228--242, 2008.

\bibitem{he2010fast}
K.~He, J.~Sun, and X.~Tang, ``Fast matting using large kernel matting laplacian
  matrices,'' in \emph{Computer Vision and Pattern Recognition (CVPR), 2010
  IEEE Conference on}.\hskip 1em plus 0.5em minus 0.4em\relax IEEE, 2010, pp.
  2165--2172.

\bibitem{xiao2014fast}
C.~Xiao, M.~Liu, D.~Xiao, Z.~Dong, and K.-L. Ma, ``Fast closed-form matting
  using a hierarchical data structure,'' \emph{Circuits and Systems for Video
  Technology, IEEE Transactions on}, vol.~24, pp. 49--62, 2014.

\bibitem{szeliski2006locally}
R.~Szeliski, ``Locally adapted hierarchical basis preconditioning,'' in
  \emph{ACM Transactions on Graphics (TOG)}, vol.~25, no.~3.\hskip 1em plus
  0.5em minus 0.4em\relax ACM, 2006, pp. 1135--1143.

\bibitem{zheng-learning}
Y.~Zheng and C.~Kambhamettu, ``{Learning Based Digital Matting},'' in
  \emph{ICCV}, 2009.

\bibitem{lee2011nonlocal}
P.~G. Lee and Y.~Wu, ``Nonlocal matting,'' in \emph{2011 IEEE Conference on
  Computer Vision and Pattern Recognition}.\hskip 1em plus 0.5em minus
  0.4em\relax IEEE, 2011, pp. 2193--2200.

\bibitem{sun2004poisson}
J.~Sun, J.~Jia, C.-K. Tang, and H.-Y. Shum, ``Poisson matting,'' in \emph{ACM
  Transactions on Graphics (ToG)}, vol.~23, no.~3.\hskip 1em plus 0.5em minus
  0.4em\relax ACM, 2004, pp. 315--321.

\bibitem{duchenne2008segmentation}
O.~Duchenne, J.~Audibert, R.~Keriven, J.~Ponce, and F.~Segonne, ``{Segmentation
  by transduction},'' in \emph{IEEE Conference on Computer Vision and Pattern
  Recognition, 2008. CVPR 2008}, 2008, pp. 1--8.

\bibitem{singaraju2009new}
D.~Singaraju, C.~Rother, and C.~Rhemann, ``{New appearance models for natural
  image matting},'' 2009.

\bibitem{mccormick1987multigrid}
S.~F. McCormick, Ed., \emph{Multigrid Methods}.\hskip 1em plus 0.5em minus
  0.4em\relax Society for Industrial and Applied Mathematics, 1987.

\bibitem{gopalakrishnan2008multigrid}
J.~Gopalakrishnan and J.~E. Pasciak, ``Multigrid convergence for second order
  elliptic problems with smooth complex coefficients,'' \emph{Computer Methods
  in Applied Mechanics and Engineering}, vol. 197, no.~49, pp. 4411--4418,
  2008.

\bibitem{pflaum2008multigrid}
C.~Pflaum, ``A multigrid conjugate gradient method,'' \emph{Applied Numerical
  Mathematics}, vol.~58, no.~12, pp. 1803--1817, 2008.

\bibitem{rhemann2009perceptually}
C.~Rhemann, C.~Rother, J.~Wang, M.~Gelautz, P.~Kohli, and P.~Rott, ``{A
  perceptually motivated online benchmark for image matting},'' in \emph{Proc.
  CVPR}.\hskip 1em plus 0.5em minus 0.4em\relax Citeseer, 2009, pp. 1826--1833.

\end{thebibliography}


\begin{IEEEbiography}[{\includegraphics[width=1in,height=1.25in]{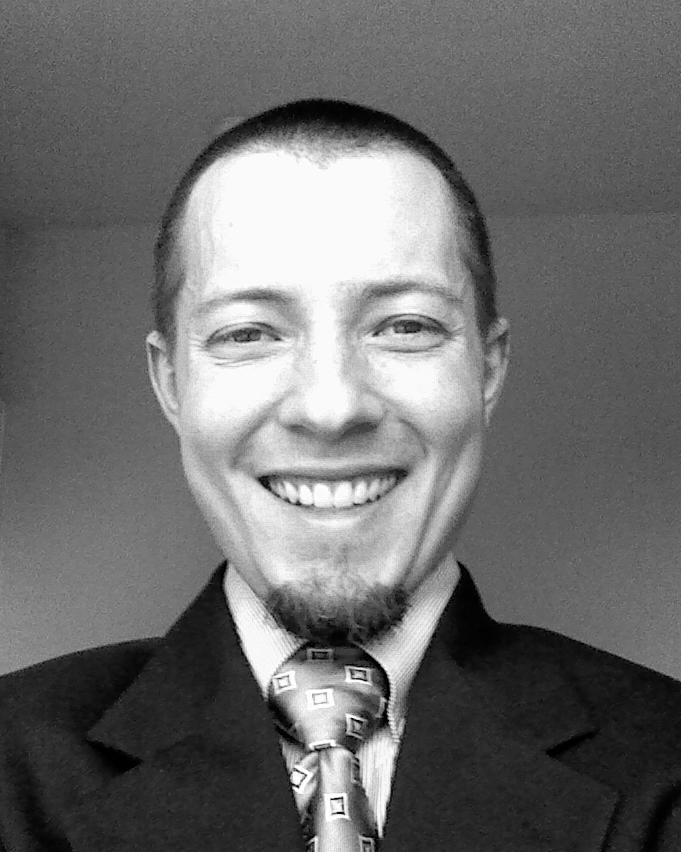}}]{Philip Greggory Lee}
 obtained his B.S. in Mathematics from Clemson University in 2008,
 where he published his first work in analytic number theory. He is currently
 pursuing a Ph.D. at Northwestern University in the Image and Video Processing
 Laboratory focusing on efficiency and
 scalability of computer vision algorithms. He is also involved with
 Northwestern's Neuroscience \& Robotics lab, working on high-speed vision
 systems for robotic manipulation.
 %{\tt http://www.linkedin.com/in/philipgreggorylee/}
\end{IEEEbiography}

\begin{IEEEbiography}[{\includegraphics[width=1in,height=1.25in]{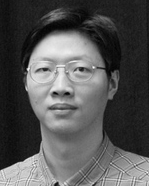}}]{Ying Wu}
 received the B.S. from Huazhong University of Science and Technology, Wuhan, China, in 1994, the M.S. from Tsinghua
 University, Beijing, China, in 1997, and the Ph.D. in electrical and computer engineering from the University of Illinois at
 Urbana-Champaign (UIUC), Urbana, Illinois, in 2001.

 From 1997 to 2001, he was a research assistant at the Beckman Institute for Advanced Science and Technology at UIUC. During summer
 1999 and 2000, he was a research intern with Microsoft Research, Redmond, Washington. In 2001, he joined the Department of Electrical
 and Computer Engineering at Northwestern University, Evanston, Illinois, as an assistant professor. He was promoted to associate
 professor in 2007 and full professor in 2012. He is currently full professor of Electrical Engineering and Computer Science at
 Northwestern University. His current research interests include computer vision, image and video analysis, pattern recognition,
 machine learning, multimedia data mining, and human-computer interaction.

 He serves as associate editors for IEEE Transactions on Pattern Aanlysis and Machine Intelligence, IEEE Transactions on Image Processing,
 IEEE Transactions on Circuits and Systems for Video Technology, SPIE Journal of Electronic Imaging, and IAPR Journal of Machine Vision and
 Applications. He received the Robert T. Chien Award at UIUC in 2001, and the NSF CAREER award in 2003. He is a senior member of the IEEE.
\end{IEEEbiography}

\end{document}